\documentclass[twocolumn,10pt]{asme2ej}

\usepackage{graphicx} 
\usepackage{caption}
\usepackage{subcaption}
\usepackage{amsmath}
\title{Grasp Control of a Cable-Driven Robotic Hand Using a PVDF Slip Detection Sensor}

\author{Negin Nikafrooz

    \affiliation{
	Terrestrial Robotics Engineering\\
	and Controls (TREC) lab\\
	Mechanical Engineering department\\
	Virginia Tech\\
    Email: nnegin71@vt.edu
    }	
}

\author{Zachary Fuge
    \affiliation{Terrestrial Robotics Engineering\\
	and Controls (TREC) lab\\
	Mechanical Engineering department\\
	Virginia Tech\\
    Email: zachf287@vt.edu
    }
}

\author{Alexander Leonessa
        \thanks{Corresponding author, Email: leonessa@vt.edu} 
    \affiliation{Terrestrial Robotics Engineering
	and Controls (TREC) lab\\
	Mechanical Engineering department\\
	Virginia Tech\\
    Email: leonessa@vt.edu
    }
}

\begin{document}
\maketitle 

\begin{abstract}
Detecting and preventing slip is a major challenge in robotic hand operation, underpinning the robot's ability to perform safe and reliable grasps. Using the robotic hand design from the authors' earlier work, a sensing and control strategy is proposed here to prevent object slippage. The robotic hand is cable-driven, single-actuated, has five fingers, and is capable of replicating most human hand motions. The slip sensing approach utilizes a piezoelectric vibration sensor, namely, polyvinylidene fluoride (PVDF), which is a flexible, thin, cheap, and highly sensitive material. The power of the filtered PVDF signal is shown to exhibit identifiable signatures during slip, thus providing a suitable slip detection mechanism. Using the PVDF feedback, an integral controller is implemented to prevent the grasped object from falling and ensure a safe, powerful, and reliable grasp. The extension movement of the robotic hand is controlled using a bend sensor, through a proportional-integral (PI) controller. The robotic hand weights 338 gr. The functionality and robustness of the proposed slip-detection sensory system and control logic implementation are evaluated through experiments.
\end{abstract}

\section{Introduction}

Tactile feedback is among the most advanced features of the human hand. It provides a variety of information about the touched object, such as, material, temperature, hardness, weight, and shape. This data can be sensed as electrical, mechanical, or thermal stimuli through the activated skin receptors and provide the required feedback to communicate with the environment and adjust the grip force and hand posture to safely grasp the object \cite{romeo2020methods}. Several research studies have been conducted to replicate the human hand grasp ability through robotic hands and grippers \cite{abd2018direction,morita2018grasping,wang2019flexible}. Tuning the grasp force and the gripper's posture are the main focus of these works, to which the environment and object uncertainties can add more challenges. For instance, applying too large grasping forces can damage a brittle object or deform a soft one, while low forces can lead to dropping the object. Therefore, a proper feedback on the status of the grasp is required to prevent the object from being damaged or dropped.

Based on the robotic hand application and the design requirements, different types of sensors have been used and developed to ensure a safe and reliable grasp. The proposed methods in the literature can be studied in two categories: controlling the grasp posture \cite{wang2011highly,yuan2017gelsight,atasoy201624,della2018toward} and monitoring the object slippage \cite{ikeda2004grip, koda2006grasping,dong2019maintaining, james2020biomimetic,wang2019flexible, abd2018direction, morita2018grasping}.

Tracking the grasp posture has been commonly utilized to perform a reliable grasp \cite{wang2011highly,yuan2017gelsight}. For instance, joint angle sensors, such as encoders, have been used to detect the shape of the grasped object \cite{wang2011highly}. Moreover, electromyography (EMG) sensors have been widely employed to classify the hand gestures for different grasp scenarios \cite{atasoy201624,della2018toward}. EMG feedback can provide accurate information about the fingers' position, grasp posture, or the object's geometry and is beneficial for performing some specific tasks, such as object manipulation. However, the robotic hand needs to be highly actuated to allow control of each finger (or a group of joints) separately and replicate the same grasp posture. The large number of actuators might make the robotic hand design bulky and unsuitable in applications where portability is required, such as prosthetic robotic hands \cite{atasoy201624,xu2016design}. Moreover, the required sensors might not be easily integrated into a compact mechanical design.

The second grasp control scheme is based on monitoring the object slippage. In the literature, this method has been studied in three phases, namely slip prediction \cite{ikeda2004grip, koda2006grasping}, incipient slip \cite{dong2019maintaining, james2020biomimetic, khamis2018papillarray, chen2018tactile}, and slip detection \cite{wang2019flexible, abd2018direction, morita2018grasping}. Slip prediction is usually implemented by monitoring the contact forces to predict when the slip is about to happen. Monitoring the friction coefficient of the contact surfaces is a commonly used, for which knowledge of both tangential and normal contact forces are required \cite{chen2018tactile}. This friction coefficient is estimated using either a multi-axial force sensor \cite{ikeda2004grip} or multiple force sensors \cite{koda2006grasping}. This approach can provide accurate information about the coefficient of friction changes and the grasp status. However, the static force sensors are usually bulky sensors and using them may result in bulky fingertip's design.

Incipient slip refers to the onset of slip where part, but not necessarily all, of the contact surfaces slip against each other \cite{chen2018tactile}. This event happens right before the whole object slips, due to the non-uniform force distribution on the contact surfaces. The customized sensors that have been developed to detect the incipient slip are mainly based on monitoring pressure distribution on the contact areas. Using an array of markers on the periphery of the contact surface to monitor the variations in pressure and coefficient of static friction is a common solution to this problem \cite{dong2019maintaining, khamis2018papillarray}. Moreover, optical sensor has been used to record the movement of the markers and detect incipient slip \cite{khamis2019novel}. Although detecting this phase of the slip can decrease the risk of dropping the grasped object, sensor integration into the robotic hand is often challenging and requires design modifications to accommodate it. Therefore, these sensors may not be applicable to all robotic hand designs.

Lastly, slip detection provides sensory feedback \textit{during the slip}. This is in contrast to the aforementioned methods that detect slip either before it happens or at the very early stages. Therefore, slip detection cannot completely eliminate slip, but is nonetheless very helpful in preventing the objects from dropping. There are customized tactile sensors in the literature that have been developed based on the idea of monitoring the distribution of a measurement (usually force or pressure) along the contact surface. Force distribution at the fingertip can provide a great amount of knowledge about shape of the object, object slippage, and the direction of slip.  For instance, BioTac is a distal phalange shape tactile sensor that has been developed to measure the force distribution on the fingertip \cite{fishel2012sensing}. This sensor has been used in the design of a robotic hand to estimate the direction of slip \cite{abd2018direction}. However, due to its mechanical design, it is not easily applicable to all robotic hand designs. A similar idea is implemented in the design of a flexible tactile sensor, in which an array of conductive rubbers has been used to detect the slip \cite{wang2019flexible}. Optic sensor \cite{yuan2017gelsight, dong2019maintaining,james2020biomimetic}, accelerometer \cite{ajoudani2014exploring}, and velocimeter \cite{morita2018grasping} are other types of sensors that have been used in the literature for slip detection.

Detecting the vibration of contact surfaces due to the slip is another method and one of the first attempts to detect the slip \cite{salisbury1967mechanical}. Among the vibration sensors, Piezoelectric materials have received more attention due to their compact size and high sensitivity. Polyvinylidene fluoride (PVDF) \cite{shirafuji2014detection} and lead zirconate titanate (PZT) \cite{rodriguez2008result} are two common types of piezoelectric material that have been used as dynamic force sensors to detect the vibration of the object slippage. These vibration sensors cannot provide an accurate measurement of the static contact forces. Therefore, in the literature, they have been mostly used along with a static sensor, such as force sensors or/and EMG devices to provide more detailed information about the contact forces and the grasp status \cite{choi2006development, rodriguez2008result}. In this study, a unique feature of the PVDF sensor is extracted and used to detect the slip of the grasped object without any need for additional static sensors. This proposed method significantly simplifies the sensor and control systems.

The main goal of this work is to develop the sensory feedback and control mechanisms that enable adaptive and reliable grasps using the slip detection technique. To this end, a robotic hand design proposed in an earlier work \cite{nikafrooz2021single} is utilized. The design is a compact, lightweight, single-actuated, and five-fingered robotic hand. The robotic hand is able to grasp objects with unknown geometries due to the novel mechanical design of the power transmission mechanism. This feature has simplified the mechanical design and the required sensory system. A PVDF sensor is used to control the flexion movement of the cable-driven robotic hand, while the extension movement is controlled using a bend sensor. A 3D printed prototype of the robotic hand is fabricated and its grasp ability is assessed through experiments.


The rest of the paper is organized as follows. The mechanical design of the robotic hand is briefly reviewed in section \ref{Desing}. The sensory system and the signal processing algorithm are discussed in section \ref{SensorSys}. The control logic is presented in section \ref{Control}, and the experimental results are discussed in section \ref{Result}. The concluding remarks are presented in section \ref{Conclusion}.

\section{Mechanical Design} \label{Desing}
\begin{figure}[t]
	\begin{center}
	\includegraphics[width=0.47\textwidth]{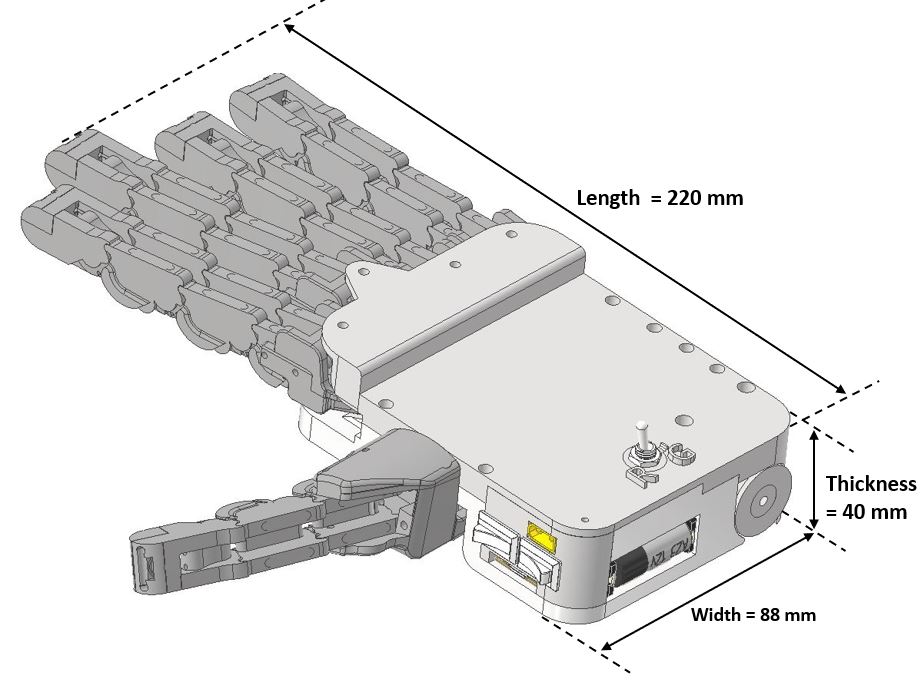}
	\caption{The CAD model of the modified robotic hand.}
	\label{Fig:RoboticHandPrototype}
	\end{center}
\end{figure}

The designed robotic hand in our previous work is modified and used in this study to evaluate the functionality of the proposed grasp control algorithm \cite{nikafrooz2021single}. The CAD model of the modified robotic hand is shown in  \ref{Fig:RoboticHandPrototype}. A brief review on the mechanical design of the robotic hand and the implemented modifications are discussed in this section. 

This five-finger and cable-driven robotic hand supports 19 (out of 20) degrees of freedom (DOF) of the human hand. The rigid external structure of the fingers and thumb is inspired from the Pisa/IIT SoftHand \cite{della2018toward}, while major modifications are implemented to improve the design, fabrication, and assembly processes. Moreover, the cable configuration is inspired from the tendon structure of the human hand anatomy. Therefore, a set of cable guides are designed along the fingers (through the phalanges) to maximize the the normal component of the contact forces (between the fingers and the object), while minimizing the effect of friction. 

A set of elastic bands are incorporated through the dorsal side of the fingers and thumb to passively help with the extension movement. The elastic bands are also responsible to keep the phalanges together, constrain the movement of phalanges, and increase the elasticity of the design. A novel cable-driven and single-actuation mechanism was designed to control the flexion movement and performing the grasp. The designed mechanism helps with performing adaptive grasps, meaning grasping objects with unknown geometries and material texture. This feature significantly helps with simplifying the sensory and control systems.

The power transmission mechanism is responsible for distributing the actuation force through the fingers and performing a powerful grasp. This mechanism includes a set of cables, which are routed through the fingers and both ends are tied to a lever. The lever pulls the cables by traveling along the palm and reduces the cable length through the fingers to perform different postures. More detailed information can be found in \cite{nikafrooz2021single}. Tie point of the cables to the lever and the elasticity of the cables are two important factors, which play an important role in performing adaptive grasps. 

The location of tie points of each cable to the lever was determined to maximize the transmitted force through the index and middle fingers (since they are involved in most of the grasp types) and minimize the force difference among all fingers. The effect of tie point positions on the power transmission mechanism was discussed in \cite{nikafrooz2021single} in detail. 

Experiments have shown that the elasticity of the cables is an important characteristic in performing adaptive grasps, as well. Therefore, a set of simulations in Adams software is conducted to evaluate the effect of the cable's elasticity on the performance of the differential mechanism \cite{msc2021a}. Through these simulations, an adaptive grasp scenario is implemented by blocking the middle finger at its initial position. Angular position of the index finger (the angular orientation of the fingertip with respect to the base of the finger, which is fixed to the palm part) and displacement of the slider along the palm part are monitored, while a constant actuation force is applied to the slider. To model different materials of the cable, Young's modulus of the cable is changed for each simulation. Young's modulus values are chosen in the range of $50-10000$ $N/mm^2$.

\begin{figure*}[ht!]
	\begin{center}
	\includegraphics[width=0.95\textwidth]{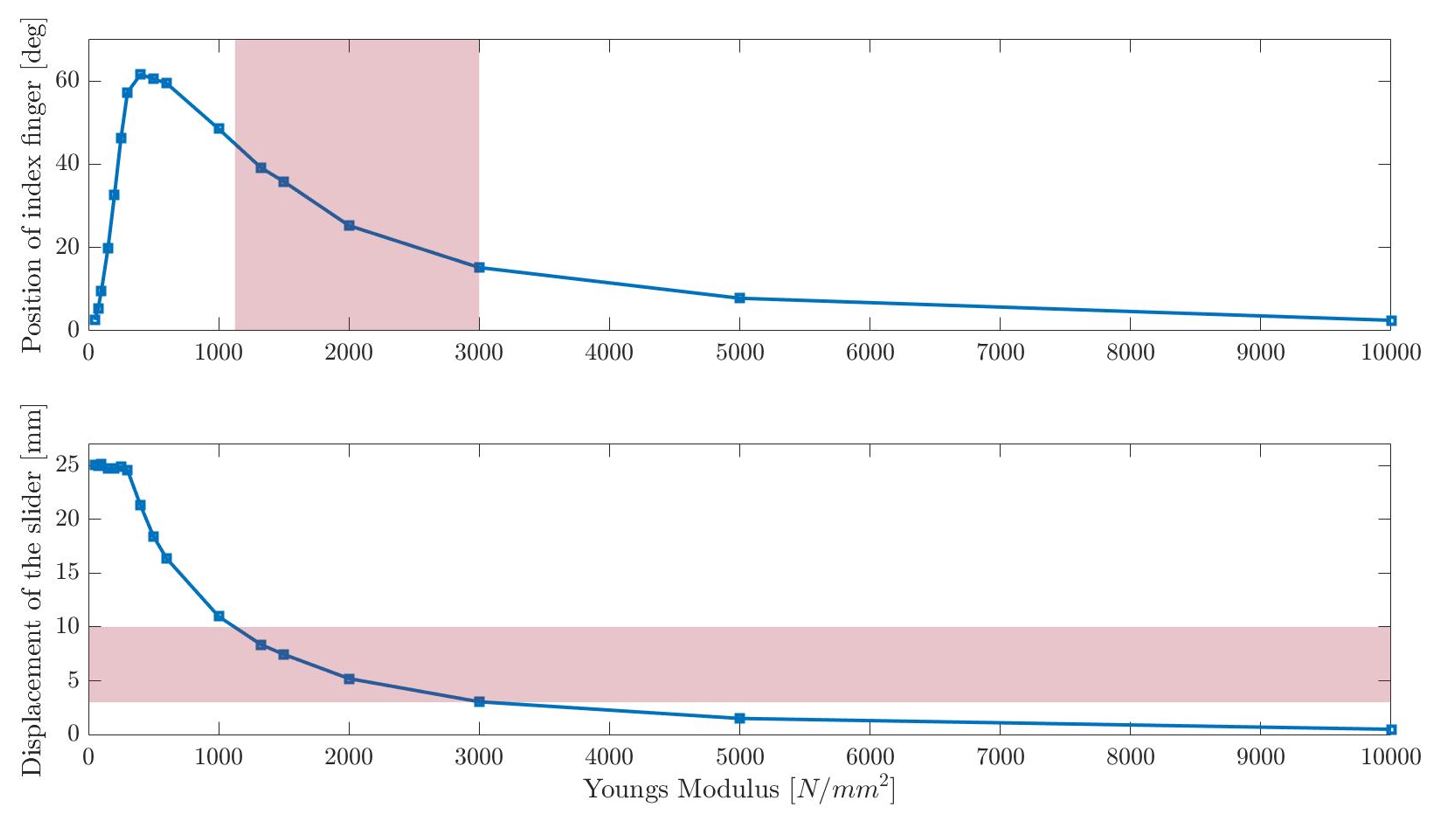}
	\caption{Evaluating the effect of cable elasticity on the performance of the differential mechanism, (top) Angular position of the index finger, (bottom) slider displacement.}
	\label{Fig:AdamsAnalysis}
	\end{center}
\end{figure*}

The result of the simulations is presented in  \ref{Fig:AdamsAnalysis}, where the top figure shows the angular position of the index finger and the bottom plot indicates the slider displacement for 17 Young's modulus values in the range of $50-10000$ $N/mm^2$. The simulations have shown that for low Young's modulus values (in the range of $50-300$ $N/mm^2$), the index finger barely bends, while the slider travels all the way through the palm (25 mm). It explains that a very elastic cable cannot transfer the actuation power properly to the fingers. For high values of Young's modulus (in the range of $3000-10000$ $N/mm^2$), the cable acts like a rigid bar. Therefore, neither the bending of the index finger nor the displacement of the slider is noticeable. 

The palm design allows for a maximum displacement of $25$ $mm$ for the slider. Based on this design constraint and the specific simulation setup, the acceptable cable elasticity is defined as that corresponding to the slider travel range of $3-10$ $mm$. This area is highlighted in the bottom part of  \ref{Fig:AdamsAnalysis}. Therefore, the acceptable cable elasticity range would be $1125-3000$ $N/mm^2$. Mapping this cable elasticity range to the top plot shows that the index finger can bend $15-47$ $deg$, when the middle finger is blocked at its initial position. Among the available cables in the market, monofilament fishing line appears to be the best choice. The Young's modulus of Nylon is reported to be in the range of $2.1-3.3$ $GPa$ \cite{matwebwebsite}, which supports the reported acceptable range of cable elasticity through the simulations. Therefore, the robotic hand is assembled using the monofilament fishing line. The result of this simulation is evaluated through experiments. This characteristic provides the opportunity of grasping unknown objects, without any need to separately control movement of each finger.

To provide a more natural appearance of the robotic hand, a minor modification is implemented on the mechanical design.
In the previous work \cite{nikafrooz2021single}, a Dynamixel motor RX-28 was used to actuate the robotic hand. In the modified design, the Dynamixel motor is replaced with a Portescap brush DC motor \cite{Portescapmotor}. The smaller size of the Portescap actuator allows it to be incorporated into the palm part. Therefore, a more compact design is provided and the overall thickness of the robotic hand is reduced by 37\% in comparison to the previous work \cite{nikafrooz2021single}.

\begin{figure*}[th!]
	\begin{center}
	\includegraphics[width=0.95\textwidth]{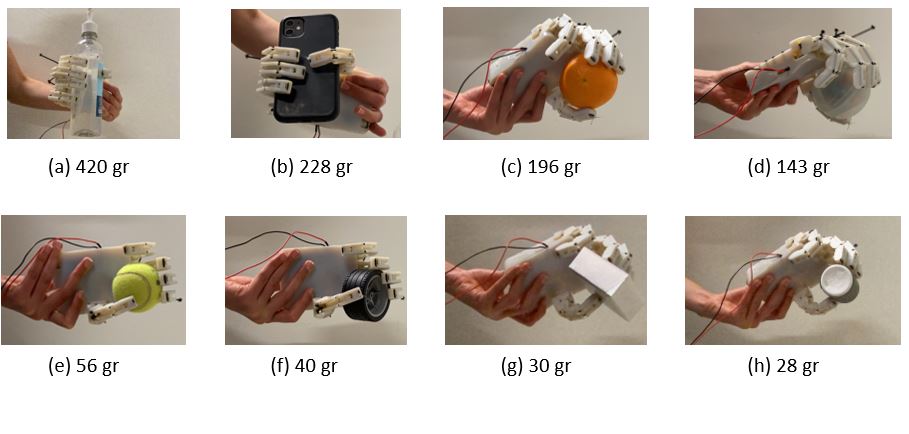}
	\caption{Assessing the adaptive grasp functionality of the robotic hand (a) sanitizer bottle, (b) iPhone 11, (c) orange, (d) plastic ball, (e) tennis ball, (f) Lego tire, (g) box, (h) Advil bottle.}
	\label{Fig:AdaptiveGrasp}
	\end{center}
\end{figure*}

A 3D printed prototype of the modified robotic hand is fabricated. The overall weight of the robotic hand is 338 gr. The adaptive grasp functionality is assessed through grasping objects with different geometries, weights, and texture. Figure \ref{Fig:AdaptiveGrasp} shows the grasp posture, where power and precision grasps are performed adaptively, based on the size and shape of the object. In the following sections, the sensor configuration, signal processing algorithm, and the implemented control logic are discussed and evaluated.

\section{Slip Detection Sensor}\label{SensorSys}
The mechanical design of the robotic hand provides the opportunity of developing a simple and compact sensory system, which is capable of performing safe and reliable grasps. In this study, the slip detection method is implemented to control the grasp movement. To detect the object slippage, a PVDF sensor is utilized. The proposed signal processing algorithm helps with extracting a novel feature to detect the object slippage and distinguish it from undesirable excitations through the experiments. The sensor structure, performance, and the signal processing implementation are discussed as follows.

\subsection{Hardware Setup}
\begin{figure}[t]
	\begin{center}
	\includegraphics[width=0.45\textwidth]{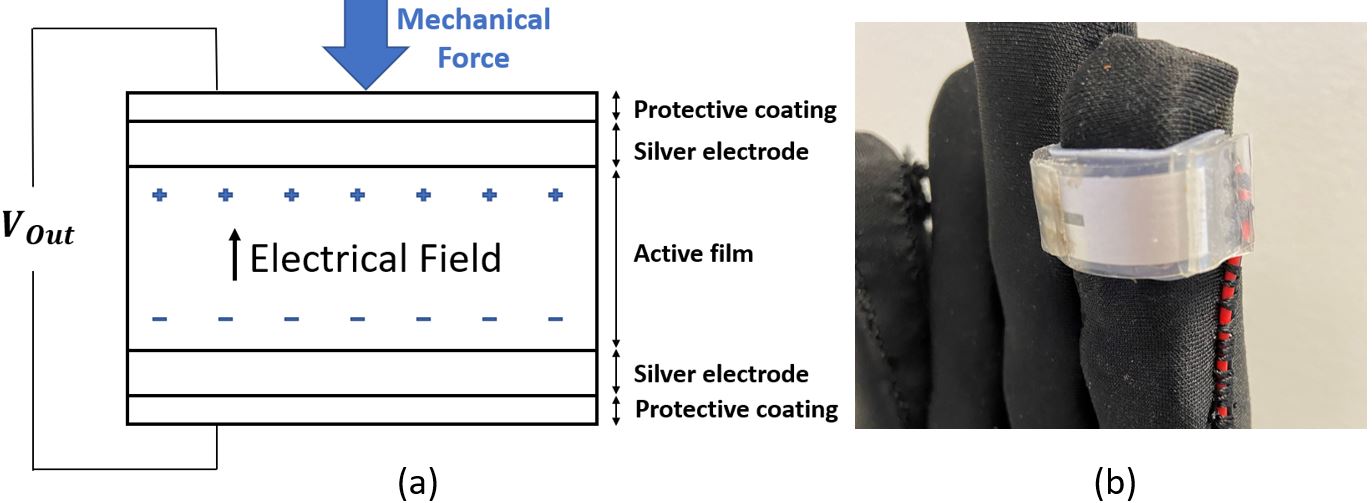}
	\caption{(a) Scheme of the PVDF sensor, and (b) sensor assembly on the index finger.}
	\label{Fig:PVDF}
	\end{center}
\end{figure}

PVDF is a flexible, thin, cheap, and highly sensitive material. As it is shown in  \ref{Fig:PVDF}(a), the sensor consists of an active film which is sandwiched between silver electrodes and protective coatings. For the sensor used in this study \cite{PolyK}, the active film thickness of the strip is 45 {\textmu}m, while the total thickness is 80-90 {\textmu}m. The small size of the sensor provides the opportunity of developing a compact sensory system.

This material exhibits piezoelectric properties, meaning that applying a mechanical force causes film deformation and consequently generates electrical charge imbalance. This property makes the PVDF film a great candidate for designing force and tactile sensors. Moreover, the PVDF film has fast response on the order of nanoseconds, which is an important characteristic for slip detection and grasp control scenarios.

PVDF is a dynamic sensor. Although the PVDF film can generate charges proportional to the applied force, after an extended period of time, the generated charge difference on the sensor electrodes decreases and eventually becomes zero. Therefore, this sensor cannot accurately detect static forces. This explains why the PVDF sensor has been used along with a static sensor, such as a resistive force sensor or EMG, in the literature \cite{choi2006development, rodriguez2008result}. Here, it is demonstrated that with a proper signal processing algorithm, a PVDF sensor can be sufficient to control the grasp performance of a robotic hand.

Figure \ref{Fig:PVDF}(b) illustrates how the sensor is incorporated into the robotic hand. Since the index finger is involved in most of the grasp postures, a PVDF strip is attached to the fingertip of this finger. The only grasp type in which the index finger is not involved is the lateral grasp. The designed robotic hand cannot perform this grasp type due to the mechanical design of the thumb \cite{nikafrooz2021single}. Therefore, assembling the PVDF sensor on the index finger does not add any restriction to the grasp performance of the robotic hand. 

The PVDF sensor is highly sensitive. Therefore, the assembly method has an important effect on the quality of the measurements. To make the assembly and maintenance processes easier, a modular sensor housing is developed. Therefore, the robotic hand is covered by a fabric glove, to which the sensors are attached. The PVDF sensor strip is assembled on a u-shaped 3D printed part, which can be easily attached to the fingertip on the robotic hand. In this study, the two ends of the sensor are fixed to the u-shaped part using double-sided tapes, while the sensors' cables are fixed along the length of the finger. Fixing the cables helps to minimize undesirable excitations. The sensing area of the sensor is the middle portion, between both fixed ends. To make sure a large enough area is in contact with the grasped object, a sufficiently large strip (10x25 mm) is chosen. This setup has been used to evaluate the proposed slip detection algorithm.

\subsection{Signal Processing}\label{SignalProcessing}
Performing a stable grasp using the slip detection method requires a combination of proper choice of sensor and signal processing algorithm. Since slip is not a physical property and cannot be directly measured, the signal processing algorithm plays an important role to help make sense of the PVDF measurements.

\begin{figure}[t]
	\begin{center}
	\includegraphics[width=0.47\textwidth]{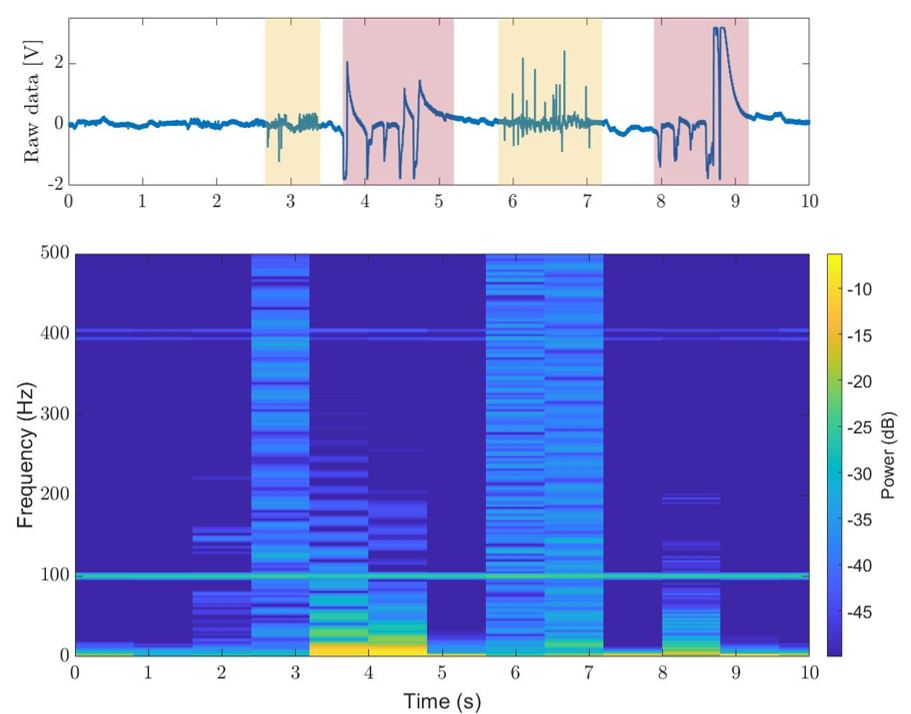}
	\caption{Collected raw data from the PVDF sensor, (Top) the PVDF signal in Volts, the highlighted bars show different implemented excitations, and (Bottom) power spectrogram analysis.}
	\label{Fig:SP1}
	\end{center}
\end{figure}

To better understand the effect of the signal processing algorithm, it is beneficial to take a look at the raw data of the PVDF sensor. Figure \ref{Fig:SP1} shows the result of an experiment, where two kinds of excitations are implemented. The PVDF sensor is excited manually by pulling, pushing, and rotating an object on its surface to emulate the object slippage scenarios. The corresponding excitations are highlighted in red. Moreover, undesirable excitations due to the cable's movement and working desk vibrations are introduced intentionally, which are highlighted in yellow.

As it can be seen in  \ref{Fig:SP1}, the undesirable excitations have as high amplitude as the mechanical excitations. Therefore, defining a simple threshold, as it is proposed in the literature \cite{shirafuji2014detection}, does not provide a robust slip detection method. The frequency content of the signal is studied using the power spectrogram analysis. Figure \ref{Fig:SP1}(Bottom) shows that both undesirable and mechanical excitations share similar frequency content. Therefore, neither the signal amplitude nor its frequency can be separately used to distinguish the different types of excitations and further investigation is required to address this challenge.

Knowledge of the PVDF characteristics can be utilized to design a simple and effective signal processing algorithm. The generated charge on the PVDF film electrodes in response to mechanical stress is governed by the piezoelectric charge constant \cite{sirohi2000fundamental}. This parameter is in the range of 22-28 pC/N for this PVDF strip, which indicates a small amount of charge generation. Therefore, an amplification is required to ensure small charges are not missed through the analysis. Moreover, it is important to mention that the applied force on the piezoelectric material electrodes is proportional to the generated charge. However, the generated charge produces current through the circuit, and then gets converted to the voltage, as it is formulated in \ref{Eq:charge}.
\begin{equation}
V_{out} = \frac{1}{C_p}\int{I dt} =  \frac{Q}{C_p}
\label{Eq:charge}
\end{equation}
In this case, the resulting signal indicates the rate of force changes. Therefore, an integrator is required to ensure the output signal truly represents the charge and consequently the applied force variations \cite{sirohi2000fundamental}.

\begin{figure}[t]
	\begin{center}
	\includegraphics[width=0.47\textwidth]{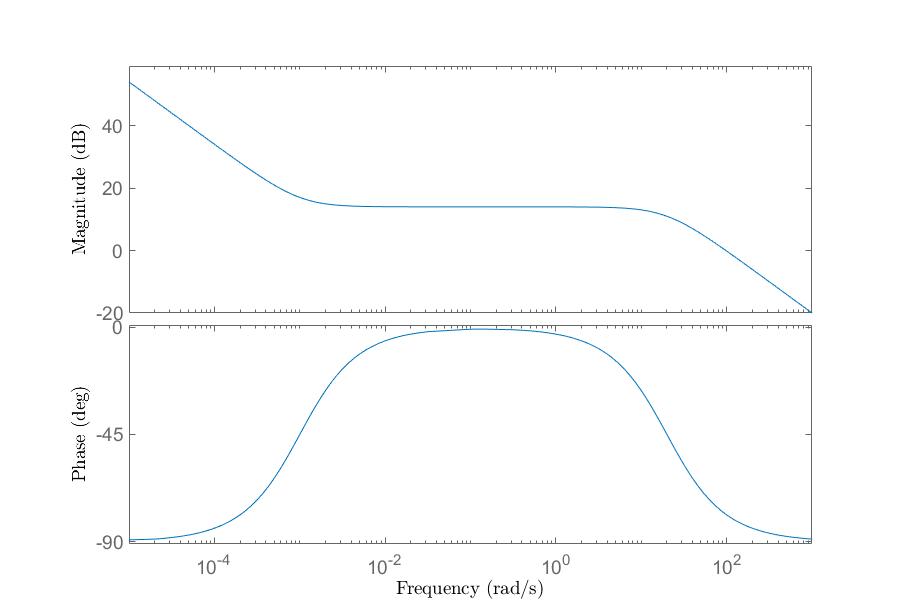}
	\caption{Bode plot of the integrator-lead compensator filter formulated in equation \ref{Eq:Filt}.}
	\label{Fig:Bodeplot}
	\end{center}
\end{figure}

Considering the discussed characteristics, an integrator-lead compensator filter is designed. The filter transfer function is given by:
\begin{equation}
G_{filt}(s) = \frac{100s+0.1}{s^2 +20S}
\label{Eq:Filt}
\end{equation}
The integrator term ensures that the output voltage represents the applied mechanical stress to the sensor electrodes, while the lead compensator term filters the high frequency noises. The Bode plot of the implemented filter can be seen in Fig. \ref{Fig:Bodeplot}.

\begin{figure}[t]
	\begin{center}
	\includegraphics[width=0.47\textwidth]{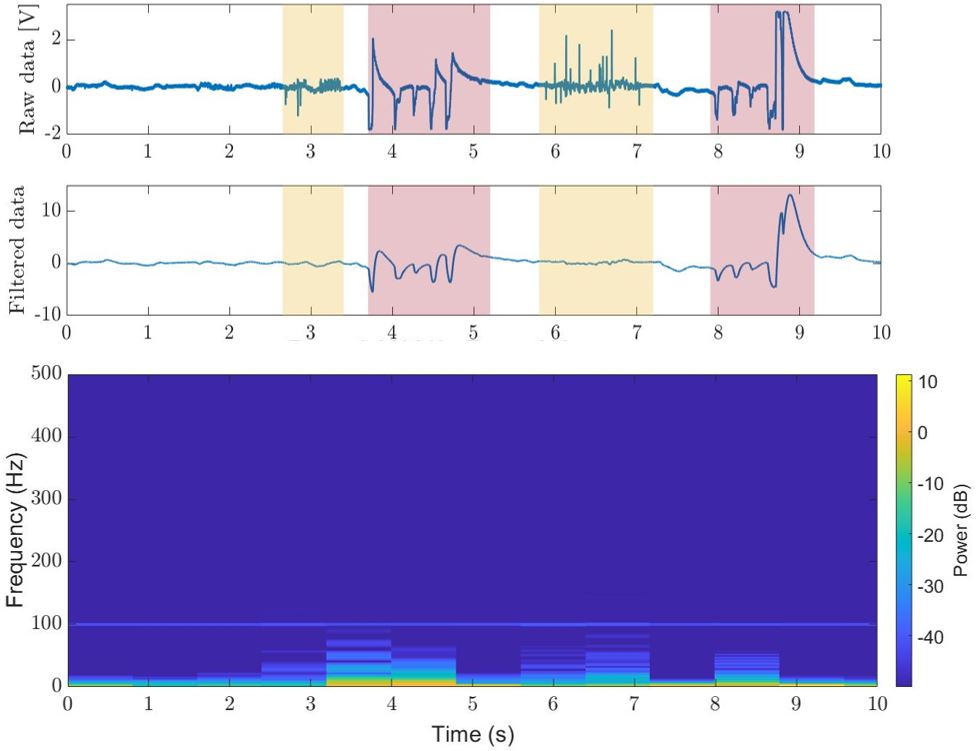}
	\caption{(Top) Raw data, (Middle) filtered signal based on the filter formulated in Eqn. \ref{Eq:Filt}, and (Bottom) the power spectrogram analysis of the filtered signal.}
	\label{Fig:filtered}
	\end{center}
\end{figure}

The filtered signal and the corresponding power spectrogram analysis are demonstrated in Fig. \ref{Fig:filtered}. It can be seen that the undesirable excitations, which are highlighted in yellow, are successfully filtered. Moreover, observing the power spectrogram analysis of the experiments shows that the signal's power due to the slip event is much higher than the undesirable excitations. Therefore, the square of the signal is used as a representation of the signal's power to provide more robust feature for the slip detection algorithm. 





\begin{figure}[t]
	\begin{center}
	\includegraphics[width=0.47\textwidth]{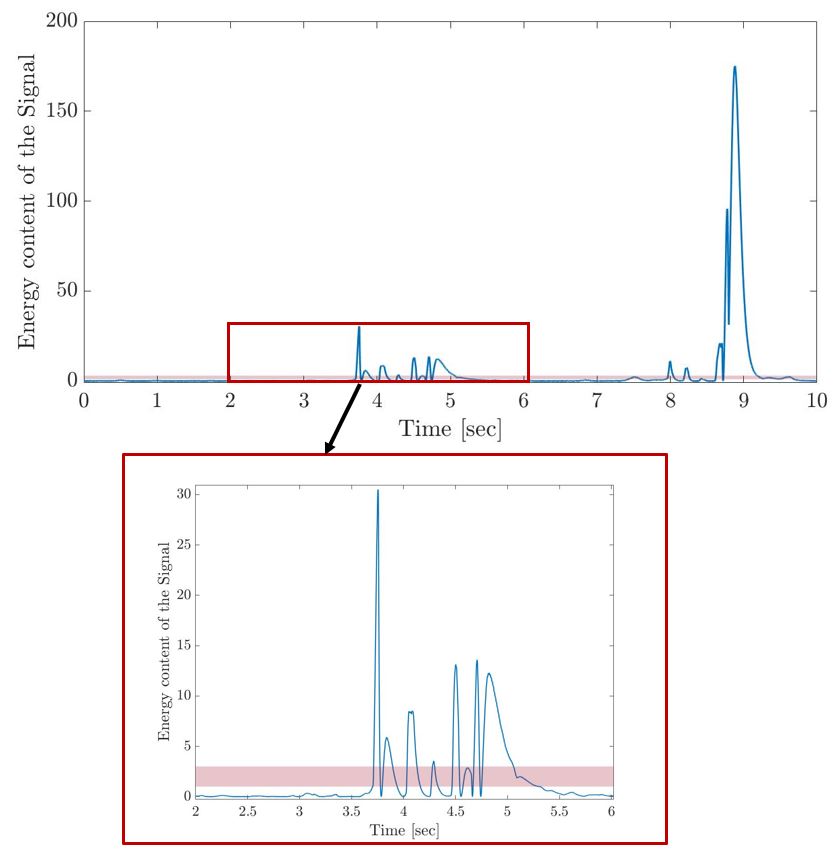}
	\caption{Deadzone threshold implementation. The highlighted band in red indicates the dead band.}
	\label{Fig:Threshold}
	\end{center}
\end{figure}

The final goal of this signal processing algorithm is to detect the slip as a binary event. Therefore, a threshold on the output voltage is required to detect the slip events. Defining a single threshold can cause actuator chatter. To address this concern, a deadzone threshold is defined, which is illustrated in Fig. \ref{Fig:Threshold} and can be formulated as follows: 
\begin{equation}
Output [k] = 
\begin{cases}
1 &  u[k]\ge u[k-1] \\
  &  and \hspace{4pt} u[k]\ge HB, \\
0 &  u[k]< u[k-1] \\
  &  and \hspace{4pt} u[k]\le LB.
\end{cases}
\label{Eq:Threshold}
\end{equation}
When the signal is increasing and passes the higher boundary (HB), the slip is detected. In this case, the actuator will be controlled accordingly to ensure a powerful and safe grasp. Moreover, when the signal is decreasing and its value gets smaller than the lower boundary (LB), the deadzone output switches to 0, which indicates the object is fully grasped or released. For this setup, the HB and LB values are tuned to be 3 and 1, respectively.

To conclude, the combination of choice of sensor, sensor assembly, and the proposed signal processing algorithm provides the opportunity of detecting the slip events. This sensory system is compact, light-weight, cheap, and reliable. It is used to control the grasp movements of the robotic hand and its functionality is evaluated through experiments, as presented in section \ref{Result}.

\section{Control Logic}\label{Control}

\begin{figure*}[th!]
	\begin{center}
	\includegraphics[width=0.97\textwidth]{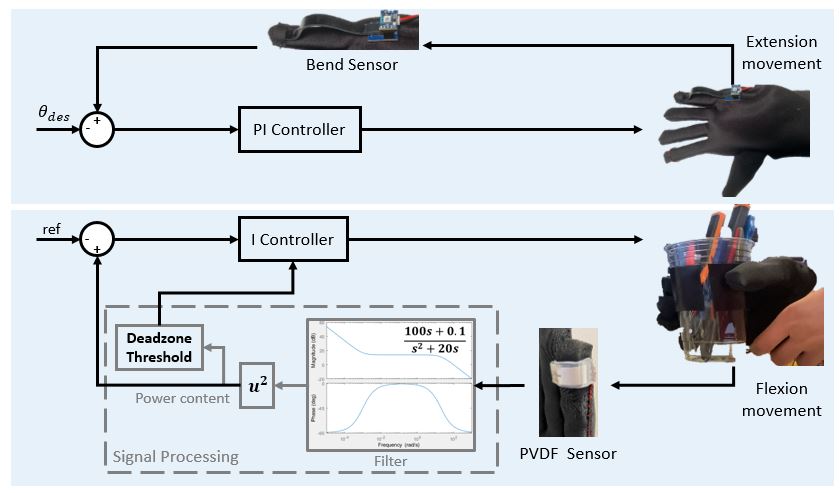}
	\caption{Control logic implementation. The extension movement is controlled using a bend sensor, while the flexion movement is controlled using the slip-detection technique. The block diagram of each of the extension and flexion movements are demonstrated in a separate blue block.}
	\label{Fig:ControlLogic}
	\end{center}
\end{figure*}

Figure \ref{Fig:ControlLogic} shows the general idea of the implemented control logic. The goal of this work is to perform safe, robust, and adaptive grasps using this robotic hand. In this study, the intention of user for performing grasp or release actions is implemented using a toggle switch.

To grasp an object, the discussed PVDF-based slip-detection algorithm is utilized. The flexion movement of the fingers starts with a constant predefined actuator's duty cycle until the sensor touches the object. When the contact is made, the main focus is on detecting the slip using the assembled PVDF sensor on the index finger and adjusting the actuator duty cycle accordingly through an integral controller (saturated at 85\%). The actuator's duty cycle is updated based on the normalized power content of the PVDF signal, when a slip event is detected through the deadzone threshold function. The reference point for the controller is zero, since that is the desired power content of the PVDF signal. The integrator is reset after the object is released.

The robotic hand is able to perform the extension movement passively through a set of elastic bands, which are incorporated into the dorsal side of the fingers and the thumb \cite{nikafrooz2021single}. Therefore, the fingers can extend even without an input from the actuator. However, this approach would lack robustness. To provide a robust, repeatable, and natural extension movements, the position feedback of the fingers is required.

The extension movement is controlled using a soft bend sensor (Bendlabs, Seattle, WA, USA) \cite{Bendlab}. The sensor output is the angular displacement measured from vectors defined by the ends of the sensor. Therefore, the bend sensor is attached to the little finger of the robotic hand by fixing both ends at the fingertip and base of the finger, as shown in Fig. \ref{Fig:ControlLogic}. The position of the fingers can be monitored using the bend sensor readings. A PI controller is implemented, where a predefined reference point (-20 degree for this setup) is used based on the initial neutral position of the fingers.

The implemented control logic is evaluated through experiments.  

\section{Results And Experimental Evaluation}\label{Result}

\begin{figure*}[th!]
	\begin{center}
	\includegraphics[width=0.85\textwidth]{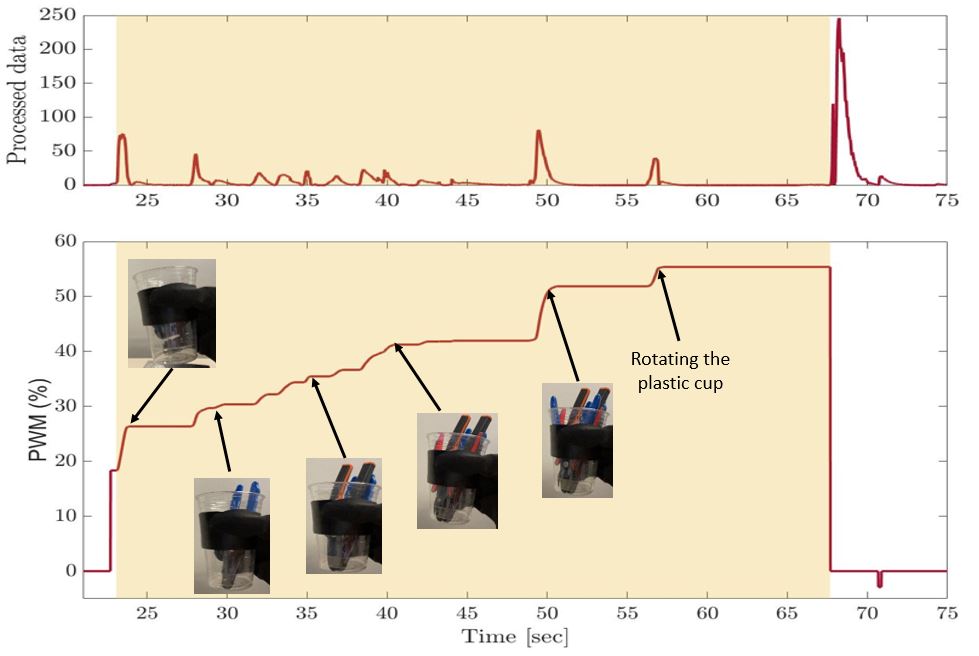}
	\caption{Grasp performance of the robotic hand, evaluating the PVDF-based slip detection method and the implemented control logic.}
	\label{Fig:Robotichandgraspctrl}
	\end{center}
\end{figure*}

To assess the functionality of the proposed sensory system, signal processing algorithm, and implemented control logic, an experiment is designed. The experiment starts with grasping  an empty plastic cup, followed by placing 4 tools in the cup. Moreover, to apply additional excitation, the plastic cup is rotated manually. The performance of the slip detection algorithm and the implemented control system is demonstrated in Fig. \ref{Fig:Robotichandgraspctrl}. 

The region highlighted in yellow indicates the time period of the experiment that the PVDF sensor is in touch with the grasped object. When the grasp mode is activated using the toggle switch, the actuator runs at a constant duty cycle until the PVDF touches the object for the first time. From that point, the actuator duty cycle is updated based on the power of the PVDF signal, using the integral controller. Processed PVDF signal can be seen in the top plot of  \ref{Fig:Robotichandgraspctrl}. Adding tools to the plastic cup causes slippage at the contact surface, which can be seen as the peaks in the top plot and the jumps in the duty cycle level. 

\begin{figure*}[th!]
	\begin{center}
	\includegraphics[width=0.91\textwidth]{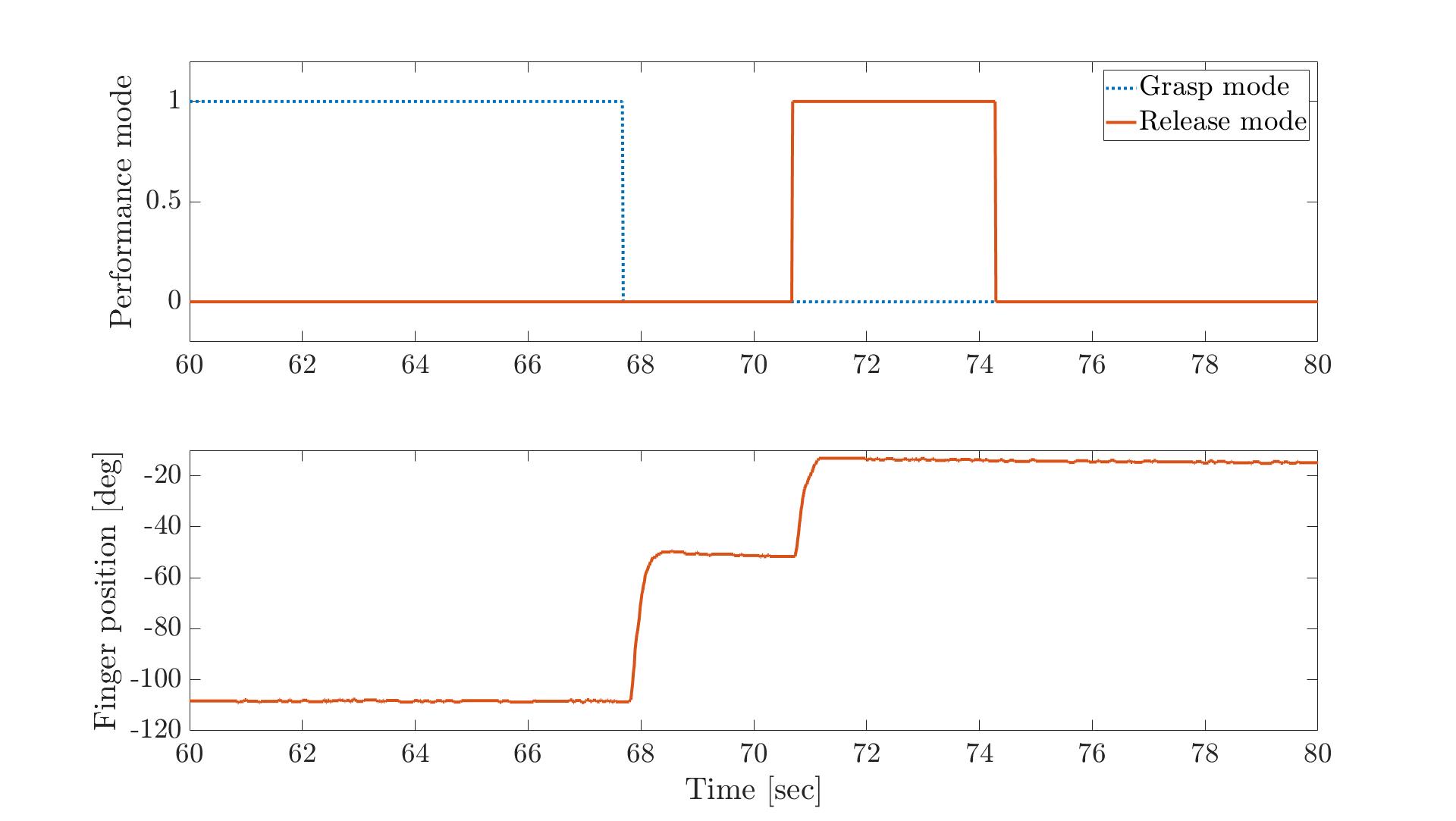}
	\caption{Release performance of the robotic hand, (Top) performance mode, which can be determined using the toggle switch, (Bottom) the bend sensor measurement. The fingers are controlled to the initial position as the release mode is activated.}
	\label{Fig:Robotichandreleasectrl}
	\end{center}
\end{figure*}

The release action can be activated using the toggle switch. The status of the toggle switch is demonstrated as the performance mode of the robotic hand in  \ref{Fig:Robotichandreleasectrl}(Top). This figure is zoomed in on the last 20 seconds of the test, where the release action is activated. The position of the little finger is shown in  \ref{Fig:Robotichandreleasectrl}(Bottom). As it can be  seen, before activating the release mode (between 68 and 70 second), the fingers are extended half way through, because of the elastic bands that are incorporated on the dorsal side of the fingers. As explained in the previous section, to perform a smooth and natural extension movement, the bend sensor is used along with the integral controller for completely extending the fingers to the initial position. To prevent the fingers' oscillation around the initial position, 8 degree deadband is considered for the error signal of the controller. The deadband's width is determined through trial and error, based on the experiments.

\section{Conclusions}\label{Conclusion}
In this work, a PVDF-based slip-detection sensory system along with the required signal processing algorithm is presented. The proposed method is used to control the grasp movement of a single-actuated and cable-driven robotic hand. Using the PVDF sensor provides the opportunity of developing a simple, reliable, compact, cheap, and lightweight sensory system. The robotic hand is controlled to perform safe and robust adaptive grasps, while performing smooth and natural extension movements. The grasp movement is controlled based on the power content of the filtered PVDF sensor readings. The Extension movement is controlled using a soft bend sensor, which is assembled on the little finger of the robotic hand. The combination of the robotic hand mechanical design and developed sensory system ensures performing reliable grasps with the robotic hand.

One of the shortcomings of the current iteration of the design is the low friction coefficient of the contact surfaces (at the fingertips). Consequently, the current robotic hand has issues grasping slippery objects. Therefore, in the experiments presented here some part of the objects are covered with rubber to increase the friction coefficient of the contact areas. This issue can be addressed in the future by using anti-slip material at the fingertips of the robotic hand.

\section*{Acknowledgment}
This material is based upon work supported by the National Science Foundation under Grants No. 1718801.

\bibliographystyle{asmems4}
\bibliography{ASMEfull}

\end{document}